# Meta-Registration: Learning Test-Time Optimization for Single-Pair Image Registration

Zachary MC Baum[1,2], Yipeng Hu[1,2], Dean C Barratt[1,2]

[1] Centre for Medical Image Computing, University College London
[2] Wellcome/EPSRC Centre for Surgical & Interventional Sciences, University College London
zachary.baum.19@ucl.ac.uk

**Abstract.** Neural networks have been proposed for medical image registration by learning, with a substantial amount of training data, the optimal transformations between image pairs. These trained networks can further be optimized on a single pair of test images - known as test-time optimization. This work formulates image registration as a meta-learning algorithm. Such networks can be trained by aligning the training image pairs while simultaneously improving test-time optimization efficacy; tasks which were previously considered two independent training and optimization processes. The proposed meta-registration is hypothesized to maximize the efficiency and effectiveness of the test-time optimization in the "outer" meta-optimization of the networks. For image guidance applications that often are time-critical yet limited in training data, the potentially gained speed and accuracy are compared with classical registration algorithms, registration networks without meta-learning, and single-pair optimization without test-time optimization data. Experiments are presented in this paper using clinical transrectal ultrasound image data from 108 prostate cancer patients. These experiments demonstrate the effectiveness of a meta-registration protocol, which yields significantly improved performance relative to existing learning-based methods. Furthermore, the meta-registration achieves comparable results to classical iterative methods in a fraction of the time, owing to its rapid test-time optimization process.

**Keywords:** Image registration. Meta-learning. Deep learning. Ultrasound.

## 1 Introduction

"Classical" pairwise approaches pose the image registration problem as an optimization for transformation, which maximizes a given image similarity measure between the transformation-warped moving image and the fixed image. Much work has been dedicated to variants in transformation models, similarity metrics and optimization algorithms [1]. These classical methods are usually applied to a single pair of images; those to be aligned. Recent learning-based methods utilize deep neural networks to predict the transformation, or simply dense displacement field (DDF), between the moving and fixed images. The networks can be optimized with a set of training pairs of images, minimizing a loss function that is based on image similarity measures or distance between corresponding segmentations [2-4]. A recent work [5] proposed to use meta-learning to adapt registration networks to new types of images, with a distinct aim of efficient domain adaptation.

More recently, deep neural networks have also been proposed to represent, or parameterize, the spatial transformation between a single pair of images. This becomes analogous to classical methods, permitting the network to be optimized "without training data"[1] [6]. It follows that the same optimization of a single pair of images may then also be applied to improve a registration network obtained from the learning-based methods – as a case of test-time optimization [7]. Both single-pair optimization approaches have shown to improve on existing methods which use learning-based registration networks alone. This may be due to the use of networks or data which are prone to overfitting, perhaps due to sensitivity to initialization, limited available training data or highly variable clinical imaging, and sometimes to underfitting due to over-constrained transformation.

Observed from these prior studies, both the single-pair methods, including those using neural networks, and the learning-based methods may have advantages in seeking pair-specific features and population-statistics-based features that are useful to align the image pair of interest. In this work, we propose using meta-learning to combine population learning and single-pair optimization, by considering image pairs in training as different meta-tasks. This allows the meta-training to optimize a meta-registration network which can be effectively and efficiently adapted to individual test image pairs, using single-pair test-time optimization.

This is particularly useful for registering ultrasound images. Ultrasound often creates challenging registration tasks with clinically acquired data given their known high variability and varying quality, due to user- and view-dependency. A rapid optimization process is also critical in enabling real-time image guidance in potentially many surgical and interventional applications. In this work, we use 3D ultrasound images obtained from transrectal ultrasound (TRUS)-guided prostate cancer interventions to demonstrate the feasibility, accuracy, and speed of optimization using the proposed meta-registration algorithm.

## 2    Methods

This section describes the proposed meta-registration using an unsupervised loss, as outlined in Fig.1. However, available segmentations of corresponding structures may readily be added for weak supervision.

---

[1] Single-pair optimization is considered as an iterative optimization in this paper to avoid confusion, as opposed to a learning-based problem where the phrase "learning without training data" may be used.

### 2.1 Unsupervised learning-based image registration

Given $N$ pairs of training moving and fixed images, $\{x_n^{mov}\}$ and $\{x_n^{fix}\}$, where $n = 1, \ldots, N$, existing approaches predict the voxel correspondence $u_n^{\omega} = f^{\omega}(x_n^{mov}, x_n^{fix})$, i.e. the transformation that aligns the two images, using a *registration network* $f^{\omega}$ with network parameters $\omega$. For an unsupervised learning algorithm, the training goal thus is minimizing a loss function over $N$ training pairs, to obtain the optimal $\omega^*$:

$$\hat{\omega} = \arg\min_{\omega} \sum_{n=1}^{N} \begin{bmatrix} \mathcal{L}_{sim}(\omega; x_n^{mov}, x_n^{fix}) + \\ \alpha^{(\omega)} \cdot \mathcal{L}_{def}(\omega; x_n^{mov}, x_n^{fix}) \end{bmatrix}, \qquad \text{Eq.(1)}$$

where $\mathcal{L}_{sim}(\omega; x_n^{mov}, x_n^{fix}) = \mathcal{L}_{sim}(x_n^{mov}(u_n^{\omega}), x_n^{fix})$ is a negative image similarity measure, a function between the transformation-warped moving images $x_n^{mov}(u_n^{\omega})$ and the fixed images $x_n^{fix}$, and $\mathcal{L}_{def}(\omega; x_n^{mov}, x_n^{fix}) = \mathcal{L}_{def}(u_n^{\omega})$ is the deformation regularization, encouraging the smoothness of the transformation $u_n^{\omega}$ weighted by a hyperparameter $\alpha^{(\omega)}$. When available, a negative weak-supervision loss based on label similarity may be added, but is omitted here for brevity.

During test time with an unseen pair of images, $x_{test}^{mov}$ and $x_{test}^{fix}$, the trained network $f^{\hat{\omega}}$ predicts the transformation that aligns the two, $u_{test}^{\hat{\omega}} = f^{\hat{\omega}}(x_{test}^{mov}, x_{test}^{fix})$.

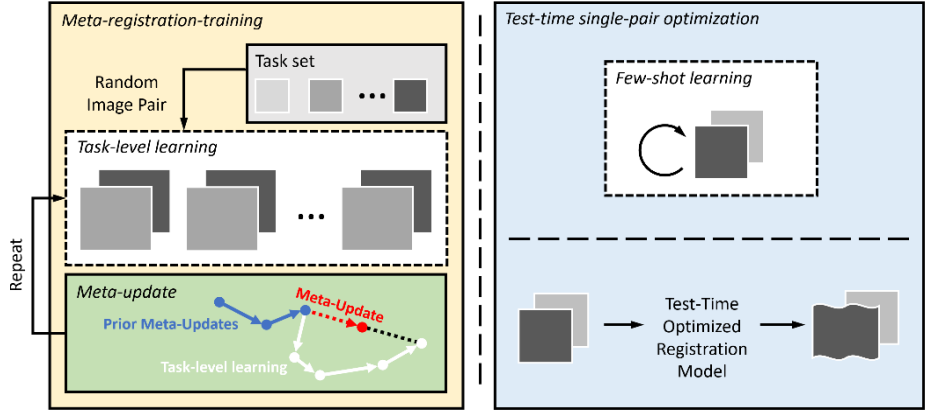

**Fig. 1.** Schematic representation of the proposed unsupervised meta-registration method for single-pair test-time optimization. A learning-based registration is trained over multiple episodes during the training phase (left). In each episode, a pair of images is sampled and repeatedly registered. Following each episode, the meta-update updates the registration model based on the learned gradients from the episode which was just completed. Once training is complete, we may optimize the registration model at test-time for a single pair of images (right) using few-shot learning to yield a registration model optimized for a specific pair of input images.

### 2.2 Test-time single-pair optimization

Now consider an optimization problem to align a pair of test images $x_{test}^{mov}$ and $x_{test}^{fix}$:

$$\hat{\theta} = \arg\min_{\theta}\big[\mathcal{L}_{sim}\big(\theta; \boldsymbol{x}_n^{mov}, \boldsymbol{x}_n^{fix}\big) + \alpha^{(\theta)} \cdot \mathcal{L}_{def}\big(\theta; \boldsymbol{x}_n^{mov}, \boldsymbol{x}_n^{fix}\big)\big], \qquad \text{Eq. (2)}$$

where $\alpha^{(\theta)}$ is the deformation hyperparameter. This is equivalent to the classical pairwise registration, iteratively optimizing a *transformation network* $\boldsymbol{f}^{\theta}$ with its randomly initialized parameters $\theta$, which (re-)parameterize the transformation $\boldsymbol{u}_n^{\theta} = \boldsymbol{f}^{\theta}\big(\boldsymbol{x}_{test}^{mov}, \boldsymbol{x}_{test}^{fix}\big)$ between $\boldsymbol{x}_{test}^{mov}$ and $\boldsymbol{x}_{test}^{fix}$.

Alternatively, when the parameters are initialized by the trained registration network parameters $\theta = \hat{\omega}$ as obtained in Eq. (1), Eq. (2) represents test-time optimization for the given test pair.

It is also noteworthy that the transformation network $\boldsymbol{f}^{\theta}$ could be a different network to the registration network $\boldsymbol{f}^{\omega}$, whilst this study uses a single network to facilitate a model-agnostic implementation of the proposed meta-registration algorithm.

### 2.3 Model-agnostic meta-learning for improving test-time optimization

This section describes the proposed meta-registration algorithm. Each pair of images is considered a different meta-task, such that a meta-training scheme can be adopted to improve the test-time optimization. During the meta-training, different meta-tasks are sampled. The resulting bi-level optimization thus becomes:

$$\omega^{*} = \arg\min_{\omega} \sum_{n=1}^{N} \begin{bmatrix} \mathcal{L}_{sim}\big(\omega; \boldsymbol{x}_n^{mov}, \boldsymbol{x}_n^{fix}, \theta^{*(n)}(\omega)\big) + \\ \alpha^{(\omega)} \cdot \mathcal{L}_{def}\big(\omega; \boldsymbol{x}_n^{mov}, \boldsymbol{x}_n^{fix}, \theta^{*(n)}(\omega)\big) \end{bmatrix}, \quad \text{Eq. (3)}$$

$$\text{s.t. } \theta^{*(n)}(\omega) = \arg\min_{\theta} \begin{bmatrix} \mathcal{L}_{sim}\big(\theta; \boldsymbol{x}_n^{mov}, \boldsymbol{x}_n^{fix}, \omega\big) + \\ \alpha^{(\theta)} \cdot \mathcal{L}_{def}\big(\theta; \boldsymbol{x}_n^{mov}, \boldsymbol{x}_n^{fix}, \omega\big) \end{bmatrix}, \qquad \text{Eq. (4)}$$

where, the outer optimization in Eq. 3 obtains the optimum meta-parameters $\omega^{*}$, such that $\theta^{*(n)}(\omega)$ is an optimized network for individual $n^{th}$ tasks. In the proposed meta-registration, $\theta$ and $\omega$ are shared network parameters. Therefore, model-agnostic meta-learning algorithms such as MAML [8] or Reptile [9] can be readily applied to solve this bi-level optimization problem.

The proposed meta-registration may be considered by two different views of combining the learning-based method and the test-time optimization: 1) it optimizes a learning-based registration network that can be used for better test-time optimization; 2) it can also be considered as an iterative method for registering a single pair of images, using a neural network to parameterize the spatial transformation, which can be initialized with prior knowledge learned from training image pairs. It is also interesting to note that data augmentation methods, commonly applied spatial variation, may be considered as the samples of individual tasks in the proposed meta-registration.

## 3 Experiments

### 3.1 The Reptile Algorithm and Meta-Registration Implementation

We adopt Reptile [9] as our gradient-based meta-learning strategy as it provides a computationally efficient optimization of the gradient-update procedure. Reptile was designed to quickly learn to perform a new task with minimal training, which suits our test-time single-pair optimization process. This is achieved in practice through a bi-level optimization. In the inner optimization loop, an episode of task-level learning is applied over $k$ mini-batches. In the outer optimization loop, Stochastic Gradient Descent is performed by using the difference between the model weights prior to and after the inner optimization loop's episode of task-level learning.

The meta-learning methodology described in this work adapts a learning-based registration method available from the unsupervised image registration framework within DeepReg, an open-source Python package for medical image registration [10]. This 'Baseline' meta-registration model architecture utilizes LocalNet [11], and was trained for 200000 iterations with the Adam optimizer [12], a mini-batch size of 4, and an initial learning rate of $1 \times 10^{-5}$. Through the meta-training phase, the value of $k$ used was 10, with an initial meta-learning rate, $\beta^{meta}$, of 0.5, linearly decaying to $1 \times 10^{-5}$ over the course of the 200,000 iterations. We utilize the sum of squared difference loss as $\mathcal{L}_{sim}$ and bending energy as $\mathcal{L}_{def}$. The deformation hyperparameter $\alpha^{(\theta)}$ was set to 10.0 to weight the deformation regularization relative to the image similarity loss. During the inner optimization, we apply data augmentation to the moving and fixed images. Each image was independently transformed by a random affine transformation, without flipping, prior to being used as input. Training required approximately 120 hours on an NVIDIA DGX-1 system using a single Tesla V100 GPU.

In the meta-test phase, we perform test-time optimization via few-shot learning with 5 gradient updates on the sampled pair of test images. This yields a test-time optimized registration model which can perform accurate registrations on the test data. In this optimization process, we use a mini-batch size of 1 and perform 5 gradient updates. Apart from these values, the few-shot learning uses the same hyperparameters as in the inner optimization loop during the meta-training phase.

### 3.2 Data

To train and evaluate the meta-registration, we used 108 intraoperative TRUS images from 76 patients, acquired during the SmartTarget clinical trials [13]. The TRUS images were split into a training set and a test set, with each comprising 88 and 20 images, respectively, where no patient appears in both sets. Images were normalized and resampled to an isotropic voxel size of $0.8 \times 0.8 \times 0.8$ mm$^3$. The TRUS segmentations of the prostate gland boundary were acquired automatically [14], and any additional landmarks were segmented manually.

### 3.3 Comparison Studies

To demonstrate the effectiveness of the meta-registration approach, we compare to a classical iterative non-rigid registration method, and two state-of-the-art architectures for deformable medical image registration [3, 11]. Additionally, we demonstrate the effects of the test-time optimization by comparing to our meta-registration baseline without any few-shot learning.

We first compare our meta-registration method with a conventional iterative registration approach, whereby stochastic gradient descent (SGD) is applied over 3000 iterations to directly learn a DDF which describes the transform between a given pair of fixed and moving images. Here, we apply a learning rate of 0.01, and use the same loss and deformation hyperparameter as in the training of our meta-registration method. We subsequently compare our meta-registration method to two widely-used approaches, LocalNet [11] and VoxelMorph [3], for deformable pairwise medical image registration using unsupervised learning. In both instances, we train the networks with identical loss, training, and deformation hyperparameters to our meta-registration method. To illustrate the effects of the test-time optimization process, as well as demonstrate the effectiveness of the meta-learned initialization, we additionally provide a comparison to only the meta-learned initialization without any test-time optimization or fine-tuning applied.

### 3.4 Evaluation of Registration Methods

The accuracy of the prostate surface registrations was quantified using the Dice similarity coefficient (DSC), and target registration error (TRE); calculated as the distance between the 3D locations of corresponding, manually identified anatomical landmarks in the TRUS images [11, 15]. Reported DSCs are computed between the transformed prostate gland label of the moving image and the ground-truth prostate gland label of the fixed image. We report TRE as the root-mean-square of the distances between landmark centroids of the pairs between the transformed moving image and the fixed image. We additionally present the computational time required at inference, on GPU, for each method.

## 4 Results and Discussion

During few-shot learning in the test-time single-pair optimization process, a gradient update and inference step require approximately 0.67s and 0.37s, respectively. Therefore, during the test-time optimization, our meta-registration method requires approximately 3.7s to be fine-tuned and provide a prediction for the specific image pair. This is notably much less than the classical method evaluated by nearly 100 times, while delivering comparable performance. Conversely, this 3.7s is nearly 10 times slower than other existing and evaluated learning-based methods which do not use any test-time optimization.

While requiring an additional 3s compared to other existing learning-based methods, the performance of DSC and TRE is improved. This performance is significantly different with respect to DSC and TRE from LocalNet and VoxelMorph, based on two-tailed paired t-tests, at a significance level of $\alpha = 0.05$.

Inference time, DSC and TRE are summarized in Table 1. Detailed results presenting DSC and TRE for each method are summarized in Figures 2 and 3. Example slices of input TRUS image pairs are provided in Figure 4 for qualitative visual assessment of the registration results for each method based on samples from the test data.

**Table 1.** Summary of DSC, TRE and computation time for our Meta-Registration method as compared to the others methods evaluated.

| Method | Time (s) Mean | DSC Mean ± STD | TRE (mm) Mean ± STD |
|---|---|---|---|
| Classical Non-Rigid | 372.36 | 0.72 ± 0.07 | 6.5 ± 2.1 |
| LocalNet [11] | 0.38 | 0.68 ± 0.09 | 7.5 ± 4.3 |
| VoxelMorph [3] | 0.39 | 0.68 ± 0.10 | 7.6 ± 4.5 |
| Meta-Registration | 0.38 | 0.73 ± 0.10 | 6.2 ± 3.8 |
| Meta-Registration with Test-Time Optimization | 3.74 | 0.74 ± 0.06 | 6.1 ± 3.7 |

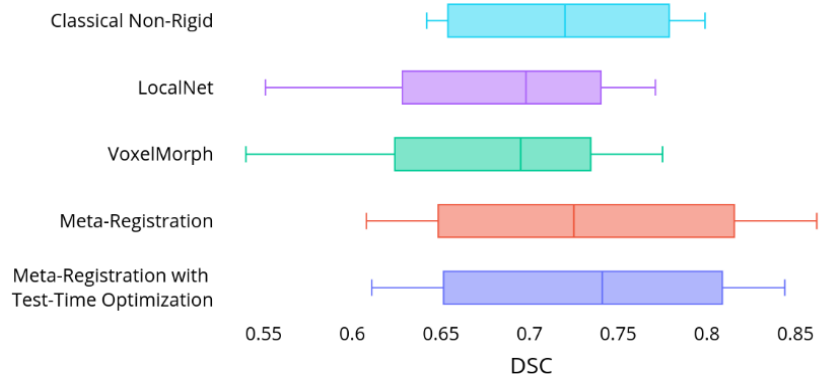

**Fig. 2.** Tukey's boxplots of DSC for all methods. Whiskers indicate 10th and 90th percentiles.

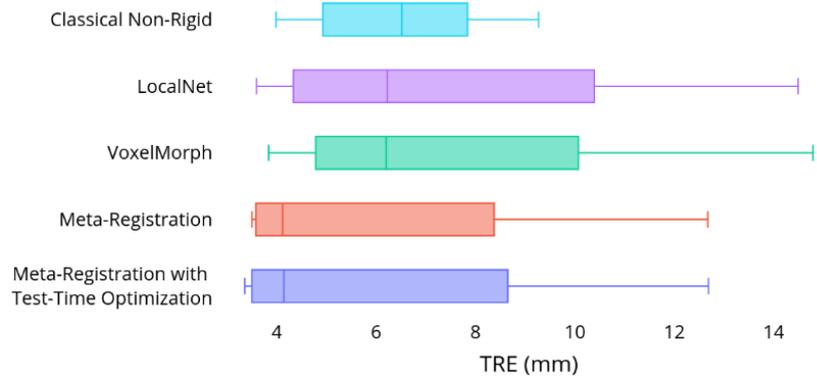

**Fig. 3.** Tukey's boxplots of TRE for all methods. Whiskers indicate 10th and 90th percentiles.

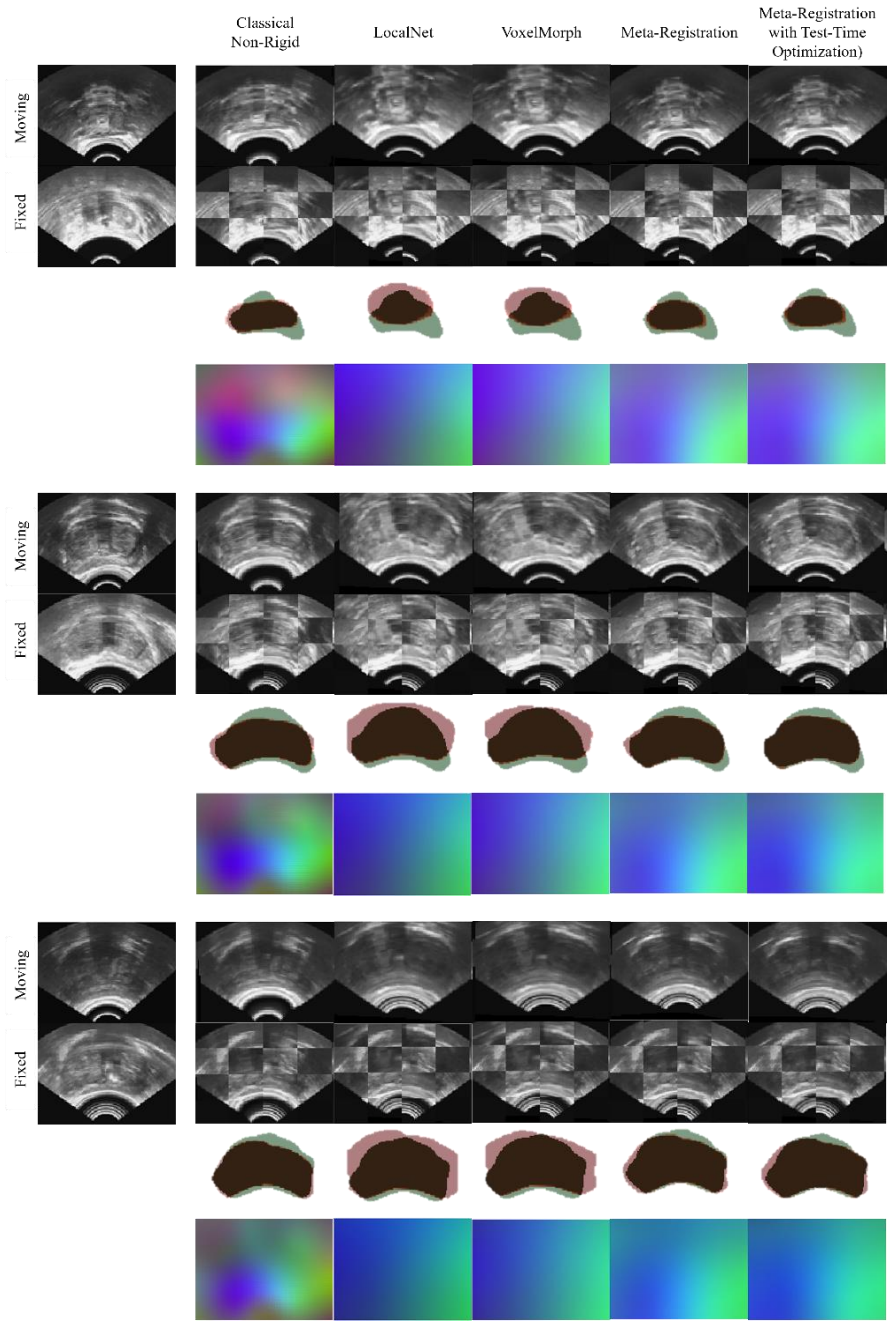

**Fig. 4.** Example image slices from one test case. The left-most column contains image slices from the fixed and moving images. Other columns present the warped image, a checkerboard of the warped and fixed images, warped prostate gland contour (Red) overlaid on the target prostate gland contour (Green), and resulting DDF using the above-labelled method.

# 5 Conclusion

We have presented a meta-registration framework for test-time single-pair optimization of ultrasound images. We obtain comparable results to time-consuming classical iterative methods in a fraction of the time, and outperform existing learning-based methods with minimal additional time required during inference for the test-time optimization process. These results demonstrate a critical step in enabling adaptive, tailored real-time image guidance in many surgical and interventional applications.

**Acknowledgments.** This work is supported by the Wellcome/EPSRC Centre for Interventional and Surgical Sciences (203145Z/16/Z). This work is also supported by the International Alliance for Cancer Early Detection, an alliance between Cancer Research UK [C28070/A30912; C73666/A31378], Canary Center at Stanford University, the University of Cambridge, OHSU Knight Cancer Institute, University College London and the University of Manchester. Z.M.C. Baum is supported by the Natural Sciences and Engineering Research Council of Canada Postgraduate Scholarships-Doctoral Program, and the University College London Overseas and Graduate Research Scholarships.